\documentclass[journal]{IEEEtran}
\usepackage{hyperref}

\ifCLASSINFOpdf
  \usepackage[pdftex]{graphicx}
\else
   \usepackage[dvips]{graphicx}
\fi
\usepackage{amsmath}
\begin{document}
%
\title{Context-Aware Multimodal Representation Learning for Spatio-Temporally Explicit Environmental Modelling}
%
%
%


\author{Julia~Peters,
        Karin~Mora,
        Miguel~D.~Mahecha,
        Chaonan~Ji,
        David~Montero,
        Clemens~Mosig,
        and~Guido~Kraemer
        \thanks{J.P., K.M., M.D.M., C.J., D.M., C.M., and G.K. are from the Environmental Data Science and Remote Sensing Group; Institute for Earth System Science and Remote Sensing, Leipzig University, Germany.}
\thanks{D.M. Montero and M.D.M. are with the German Centre for Integrative Biodiversity Research (iDiv) Halle--Jena--Leipzig, Puschstraße 4, 04103 Leipzig, Germany.}%
\thanks{Corresponding author: Julia Peters (e-mail: julia.peters@uni-leipzig.de).}        
        }




%

%
%

\markboth{IEEE TRANSACTIONS ON GEOSCIENCE AND REMOTE SENSING, VOL. XY, 202X}%
{Shell \MakeLowercase{\textit{et al.}}: Bare Demo of IEEEtran.cls for IEEE Journals}
%



\maketitle

\begin{abstract}
Earth observation (EO) foundation models have emerged as an effective approach to derive latent representations of the Earth system from various remote sensing sensors. These models produce embeddings that can be used as analysis-ready datasets, enabling the modelling of ecosystem dynamics without extensive sensor-specific preprocessing. However, existing models typically operate at large spatial or temporal contexts, limiting their use for ecological analyses that require both fine spatial detail and high temporal fidelity. To overcome these limitations, we propose a representation learning framework that integrates different EO modalities into a unified feature space at high spatio-temporal resolution. We introduce the framework using \mbox{Sentinel-1} and \mbox{Sentinel-2} data as representative modalities. Our approach produces a latent space at native 10 m resolution and the temporal frequency of cloud-free \mbox{Sentinel-2} acquisitions. Each sensor is first modeled independently to capture its sensor-specific characteristics. Their representations are then combined into a shared model. This two-stage design enables modality-specific optimisation and easy extension to new sensors, retaining pretrained encoders while retraining only fusion layers. This enables the model to capture complementary remote sensing data and to preserve coherence across space and time. Qualitative analyses reveal that the learned embeddings exhibit high spatial and semantic consistency across heterogeneous landscapes. Quantitative evaluation in modelling Gross Primary Production reveals that they encode ecologically meaningful patterns and retain sufficient temporal fidelity to support fine-scale analyses. Overall, the proposed framework provides a flexible, analysis-ready representation learning approach for environmental applications requiring diverse spatial and temporal resolutions.

\end{abstract}


{\itshape \small
This work has been submitted to the IEEE for possible publication. 
Copyright may be transferred without notice, after which this version may no longer be accessible.
}

%
\IEEEpeerreviewmaketitle

\section{Introduction}
\IEEEPARstart{R}{ecent} years have seen the rise of foundation models across domains. Through large-scale pretraining, these models produce embeddings, compact representations that capture essential structures in high-dimensional data and can be transferred to diverse downstream tasks. In language and vision, early models such as Word2Vec \cite{mikolov2013efficient} showed how embeddings capture semantic relationships, paving the way for large-scale models such as GPT \cite{mann2020language} and CLIP \cite{radford2021learning} that generalize flexibly across diverse tasks and modalities. In biology, AlphaFold2 \cite{jumper2020alphafold} demonstrates how embeddings can encode structural constraints of proteins, transforming the way molecular function is understood \cite{sindeeva2025aftoolkit, nguyen2025unified}. Similar principles extend to speech and audio, where models such as Wav2Vec \cite{schneider2019wav2vec} learn latent representations that support transcription or speaker identification, and to medicine, where embeddings derived from clinical text or imaging enable diagnostic support and patient similarity analyses \cite{habibi2017deep, yu2019biobert}.

Over the past decades, a growing number of Earth observation (EO) missions have generated vast and diverse archives of remote sensing data, spanning multiple spatial, spectral, and temporal scales \cite{ma2015remote, montero2024earth}. These datasets provide a comprehensive view of environmental dynamics, from local ecosystem states to global vegetation trends. Satellite missions such as \mbox{Sentinel-1} and \mbox{Sentinel-2} deliver observations at high spatial and temporal resolution, enabling detailed assessment of land surface conditions and vegetation structure.

Despite this large-scale data availability, transforming satellite acquisitions into meaningful and transferable insights remains challenging. Remote sensing data is inherently heterogeneous: observations differ across sensors, acquisition geometries, and atmospheric conditions, and therefore must be treated individually. Optical and radar missions, for instance, require distinct preprocessing steps, ranging from atmospheric correction to speckle filtering. In addition, sensors operate on different revisit schedules, producing asynchronous time series that require careful temporal alignment before a joint analysis. These factors complicate the construction of consistent and comparable datasets across space, time, and modality when analysing land surface processes.


EO foundation models have recently emerged as a promising solution \cite{longepe2025earth, galaz2025ai}. Overall, these models aim to learn generalized representations of the Earth system across multiple sensors, regions, and time periods that can be directly applied to diverse downstream tasks.
They produce transferable, analysis-ready embeddings, bypassing the need for expert-driven preprocessing and harmonization across sensors.
Such embeddings have demonstrated the potential to capture ecosystem dynamics at scale for various modalities, e.g.,  \cite{fibaek2024phileo, allen2024m3leo}. However, they typically operate at fixed spatio-temporal granularities that limit analytical flexibility. 
Google’s AlphaEarth~\cite{brown2025alphaearth} embeddings provide consistent large-scale representations but operate on yearly intervals that do not resolve the intra-annual variability essential for downstream applications, such as analysing fast-changing vegetation dynamics~\cite{wen2025detection} or capturing phenological processes~\cite{simmonds2019cue, shutt2019environmental}.
Similarly, TESSERA \cite{feng2025tessera} embeddings compress \mbox{Sentinel-1} and \mbox{Sentinel-2} time series into a single annual representation per pixel, offering per-pixel coverage but limited temporal flexibility. 
Conversely, temporally resolved approaches such as the Copernicus Foundation Model \cite{wang2025towards} and Major TOM \cite{czerkawski2024global} operate at large spatial contexts ($26 \times 26~\text{km}^2$ and $10 \times 10~\text{km}^2$, respectively), limiting their utility for fine-grained biodiversity assessments \cite{robertson2023effects}, or habitat mapping \cite{gregovich2025vegetation, de2021identifying} 
where local heterogeneity is crucial. 
While these embeddings demonstrate high scalability, their spatial and temporal constraints limit their ability to capture local, short-term vegetation dynamics or to match the specific spatial or temporal horizons required by ecological tasks. Consequently, their applicability for ecological analyses remains restricted.

To address these limitations, we propose a multimodal, context-aware representation learning framework that integrates heterogeneous EO modalities into a unified latent feature space at high spatio-temporal resolution. 
In this work, we demonstrate the framework using \mbox{Sentinel-1} radar and \mbox{Sentinel-2} optical data as representative modalities. 
The approach enables to encode complementary structural and spectral information while preserving coherence across space and time.
By maintaining the native 10 m spatial resolution and the temporal fidelity of cloud-free \mbox{Sentinel-2} observations, it provides compact, spatio-temporally explicit descriptors of land-surface conditions that can be flexibly applied to diverse Earth system applications.

To demonstrate that the learned embeddings constitute a general-purpose, analysis-ready representation for environmental modelling, we performed two complementary assessments.
First, a spatial assessment based on principal component analysis (PCA) compares internal feature projections with corresponding \mbox{Sentinel-2} RGB imagery to examine spatial coherence and semantic consistency across heterogeneous landscapes.
Second, a temporal assessment uses our embeddings as predictors in a Gross Primary Production (GPP) modelling task to test whether they encode ecologically meaningful temporal dynamics related to vegetation productivity.

\section{Data}

To ensure a diverse representation of European ecosystems, we randomly selected 250 training cube locations of $900~\text{m} \times 900~\text{m}$ across different land cover classes using the  European Space Agency (ESA) Climate Change Initiative (CCI) land cover map \cite{esa_landcover_2024}. The sampling domain was restricted to Central and Western Europe, specifically the bounding box $[0^\circ, 42^\circ, 30^\circ, 62^\circ]$ (west, south, east, north). The distribution of selected land cover types is summarized in Table~\ref{tab:landcover}. We maintained a minimum spatial separation of 50~km among cube centers to reduce spatial autocorrelation.

At each selected location, we extracted two annual \mbox{Sentinel-2}
 Level-2A \cite{drusch2012sentinel} data cubes from acquisition periods separated by at least one year. 
For model training, we use the ten \mbox{Sentinel-2} spectral bands at 10 and 20\,m spatial resolution were retained (B01, B02, B03, B04, B05, B06, B07, B08, B8A, B11, B12). Each cube covers a continuous 12-month period between 2017 and 2024. This temporal block sampling reduces temporal biases and ensures coverage across different seasonal and inter-annual conditions. The temporal resolution follows the \mbox{Sentinel-2} revisit cycle, yielding approximately 130–150 observations per year in Central Europe.

We applied two preprocessing steps across all cubes. First, we performed a nadir BRDF correction using view geometry parameters and the \textit{sen2nbar} \cite{montero2024facilitating} Python library. Second, we masked cloud and cloud shadows using an AI-based model from CloudSEN12 \cite{aybar2022cloudsen12}.

We retrieved the corresponding spatially and temporally aligned \mbox{Sentinel-1} data from Microsoft Planetary Computer\footnote{\url{https://planetarycomputer.microsoft.com/dataset/Sentinel-1-rtc}}. 
This collection provides radiometrically terrain-corrected (RTC) radar data~\cite{small2011flattening}, which uses PlanetDEM\footnote{\url{https://www.planetobserver.com/global-elevation-data}} as the elevation reference for normalization. 
To further reduce speckle noise and short-term fluctuations caused by rainfall or varying sensor geometry, we applied a three-frame rolling mean to the RTC time series in the linear scale.
Finally, we normalised Sentinel-1 data and Sentinel-2 reflectance to the same 0–1 range.

To prevent data leakage, all samples derived from a single cube were used exclusively for either training, validation, or testing.
The full dataset (composed of date from both sensors) was divided into a training set (375 cubes, 75\%), a validation set (83 cubes, 17\%), and a test set (42 cubes, 8\%) to ensure a reliable model evaluation.

\begin{table}[!t]
\caption{Distribution of land cover classes within the first set of training cubes.}
\label{tab:landcover}
\centering
\begin{tabular}{l c}
\hline
\textbf{Land cover class} & \textbf{Distribution} \\
\hline
Needle-leaved forest & 20\% \\
Broadleaved forest & 20\% \\
Grassland & 20\% \\
Urban & 5\% \\
Others (without water) & 35\% \\
\hline
\end{tabular}
\end{table}

\section{Methods}
Learning spatio-temporally explicit EO representations at high resolution requires integrating heterogeneous sensor modalities. Different EO missions capture complementary but fundamentally distinct physical signals, such as radar from \mbox{Sentinel-1} and multispectral reflectance from \mbox{Sentinel-2}, which vary strongly in dynamic range, noise characteristics, and spatio-temporal sensitivity~\cite{conti2025advancing}. Because each modality tends to generalise under different conditions, end-to-end joint training with a single optimisation strategy is often sub-optimal~\cite{wang2020makes, sun2021learning, yao2022modality}. Consequently, multimodal learning must reconcile modality-specific structures with cross-modal relationships while maintaining stable optimisation and coherent representations.

We adopt a staged learning strategy in which each modality is first modeled independently through a dedicated autoencoder~\cite{yu2017multi,yao2022modality, yang2023code}. This unimodal pretraining phase enables the network to capture intrinsic spatio-temporal structures specific to each sensor, yielding stable and well-structured latent representations. Once pretrained, the modality-specific layers are frozen, and additional lightweight fusion layers are introduced to learn the interactions between modalities. During this second phase, only the newly added fusion components are trained, ensuring efficient training while preserving the integrity of the unimodal representations.

During this staged approach, we use a context-aware learning strategy, in which the reconstruction objective emphasizes the central coordinate while simultaneously incorporating information from its spatial and temporal neighborhood. This enables the model to incorporate dependencies among nearby observations, allowing the model to use surrounding information as additional cues.

The following subsections detail the core elements of this framework.

\begin{figure}[t]
  \centering
  \includegraphics[width=0.75\linewidth]{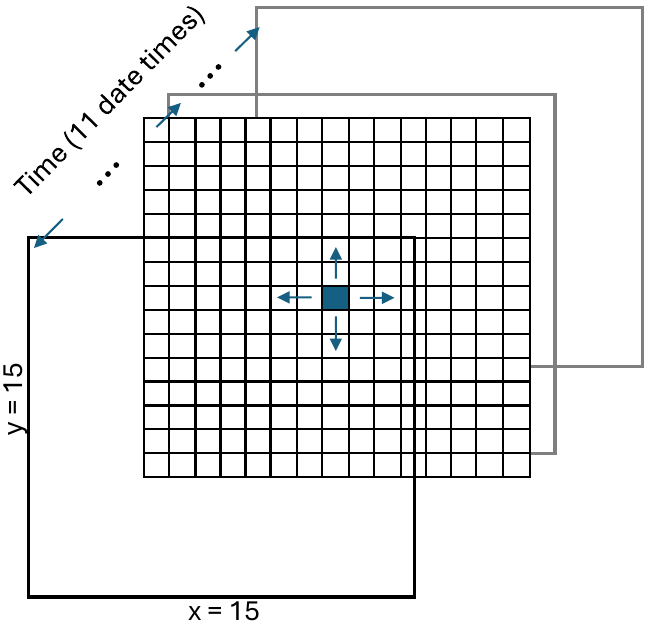}
  \caption{Spatio-temporal context window used for reconstruction. The autoencoder processes 15×15 pixel patches over 11 time steps, embedding each observation in its environmental context. The central pixel is reconstructed, while neighboring pixels provide context with exponentially decaying loss weights by distance.}
  \label{fig:context-aware-MAE}
\end{figure}

\begin{figure*}[ht]  
  \centering
  \includegraphics[page=1,width=\textwidth]{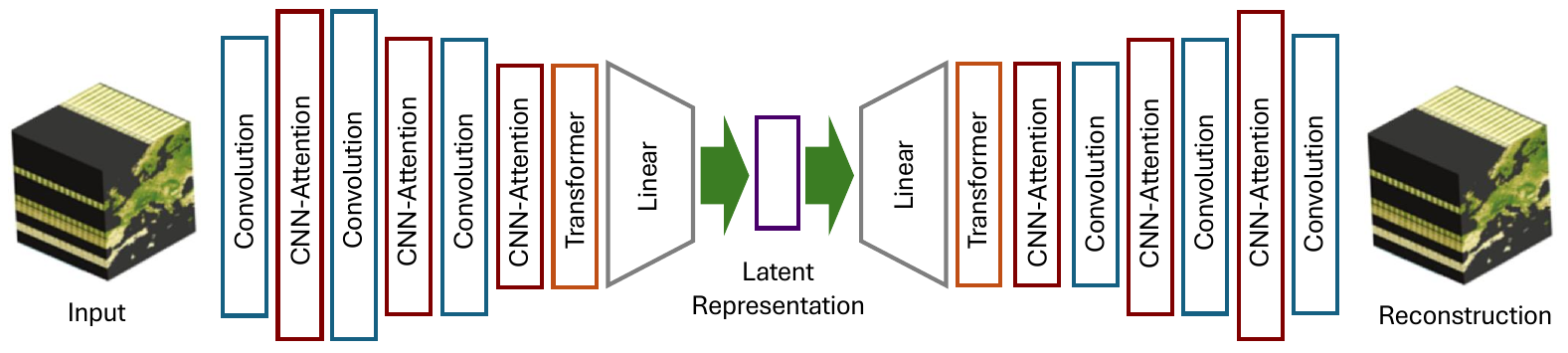}
  \caption{Overview of the Modality Autoencoder. The model processes spatio-temporal input patches of size $15 \times 15$ pixels over 11 temporal frames, using either 10 \mbox{Sentinel-2} bands or 2 \mbox{Sentinel-1} bands. It integrates convolutional layers, CNN-Attention modules (Fig.~\ref{fig:multiscale_block}), transformer-based temporal encoders, and linear layers. At the bottleneck, the spectral information is compressed into a latent representation, from which the decoder reconstructs the original input.}
  \label{fig:modality-encoder}
\end{figure*}

\subsection{Context-Aware Learning} \label{context-aware-learning}

Context-aware learning is a well-established strategy in machine learning that enhances representation learning by explicitly considering the relationships among neighboring observations in space and time. Rather than treating each input independently, it allows the model to leverage contextual cues from surrounding data points to infer more coherent and meaningful representations. This concept has proven effective across multiple domains, including computer vision~\cite{byeon2018contextvp, nascimento2018context}, natural language processing~\cite{maheshwary2021context, yang2021context}, and geospatial modelling~\cite{zhang2019cad, song2024context}, where local dependencies strongly influence the target signal. 

Our model adopts a context-aware training strategy that emphasizes the central spatio-temporal coordinate and softly incorporates its neighborhood (Fig.~\ref{fig:context-aware-MAE}). To achieve this, the autoencoder processes spatio-temporal patches of $15 \times 15$ pixels across 11 time steps. The central pixel serves as the reconstruction target, while its surroundings provide contextual cues.
Observations that are spatially or temporally closer receive higher weighting during training, reflecting their stronger ecological relevance. This encourages the model to produce features that are locally specific yet contextually informed, aiming to improve robustness to noise and spatio-temporal irregularities.

To reflect the decreasing relevance of distant observations, pixel-wise loss weights decay exponentially with spatial and temporal distance from the center, defined as
\begin{equation}
\mathcal{L}_{\text{MAE}} = \frac{1}{\sum_{i,j,t} w_{i,j,t}} \sum_{i,j,t} w_{i,j,t} \, \left| \hat{x}_{i,j,t} - x_{i,j,t} \right|,
\end{equation}
with $w_{i,j,t} = \alpha^{d_{i,j,t}}$, $\alpha \in (0,1)$. Here, $d_{i,j,t }$ denotes the distance from the central point of the spatio-temporal sample.
In our experiments, we set $\alpha = 0.1$ to obtain a steep exponential decay that prioritises information from the immediate spatial and temporal neighborhood but still retain broader contextual signals. 

For \mbox{Sentinel-1}, this formulation effectively captured spatial and temporal structures, emphasizing the central coordinate, leveraging contextual information from surrounding pixels. For \mbox{Sentinel-2}, however, the reconstruction quality of the central point within the spatio-temporal sample improved substantially (by 42\%) when the loss formulation was refined. The central point, which we primarily aim to reconstruct accurately in a context-aware manner, benefited from incorporating additional perceptual and spectral consistency terms.

The Structural Similarity Index Measure \cite{wang2004image} (SSIM) improves the perceptual and structural fidelity of the reconstruction by optimizing spatial coherence in brightness, contrast, and local patterns. It is defined as:
\begin{equation}
\text{SSIM}(x, \hat{x}) = 
\frac{(2\mu_x \mu_{\hat{x}} + c_1)(2\sigma_{x\hat{x}} + c_2)}
{(\mu_x^2 + \mu_{\hat{x}}^2 + c_1)(\sigma_x^2 + \sigma_{\hat{x}}^2 + c_2)},
\end{equation}
where $\mu_x$ and $\mu_{\hat{x}}$ are the local means, $\sigma_x$ and $\sigma_{\hat{x}}$ the variances, and $\sigma_{x\hat{x}}$ the covariance between the original and reconstructed patches. Constants $c_1$ and $c_2$ stabilize the division in regions of low variance.
In the loss formulation, SSIM is converted into a discrepancy measure:
\begin{equation}
\mathcal{L}_{\text{SSIM}} = 1 - \text{SSIM}(x, \hat{x}).
\end{equation}

The Spectral Angle Mapper \cite{kruse1993spectral} (SAM) enforces spectral consistency by minimizing the angular difference between the original and reconstructed spectral vectors. It is defined as:
\begin{equation}
\mathcal{L}_{\text{SAM}} = 
\arccos \left(
\frac{\hat{x} \cdot x}{\|\hat{x}\| \, \|x\|}
\right),
\end{equation}
where $\hat{x}$ and $x$ denote the reconstructed and original spectral vectors, respectively. 

The resulting \textit{hybrid loss} designed to learn \mbox{Sentinel-2} data integrates three complementary components:

\begin{equation}
\mathcal{L}_{\text{total}} = 0.33\,\mathcal{L}_{\text{MAE}} + 0.02\,\mathcal{L}_{\text{SSIM}} + 0.65\,\mathcal{L}_{\text{SAM}}.
\end{equation}

Here, $\mathcal{L}_{\text{SSIM}}$ maintains spatial structure, and $\mathcal{L}_{\text{SAM}}$ enforces spectral consistency at the central reconstruction target. The weighting factors were determined empirically based on reconstruction performance of the central pixel of the spatio-temporal sample. This configuration balances spatial context and spectral precision, improving reconstruction quality for \mbox{Sentinel-2}.

\subsection{Modality Pretraining}
\label{modality-pretraining}

In the first training stage, modality-specific autoencoders were trained separately for \mbox{Sentinel-1} and \mbox{Sentinel-2} to capture the structures of radar and optical observations.
From the preprocessed data cubes, fixed-size patches of $15 \times 15$ pixels across 11 time steps were extracted to provide each model with local and contextual information. For \mbox{Sentinel-1} sequences of 40 timestamps were sampled using strides of 40 in time and 15 in space. From these 40 acquisitions 11 frames were randomly selected to enhance temporal generalization. In contrast, the lower availability of cloud-free \mbox{Sentinel-2} data required shorter sequences of 20 frames, extracted with a temporal stride of 17 and a spatial stride of 9. Again, 11 frames were again randomly chosen per sample, and only the ten optical bands at 10–20 m resolution were retained for this process.

\begin{figure}[t]
  \centering
    \includegraphics[width=0.85\linewidth]{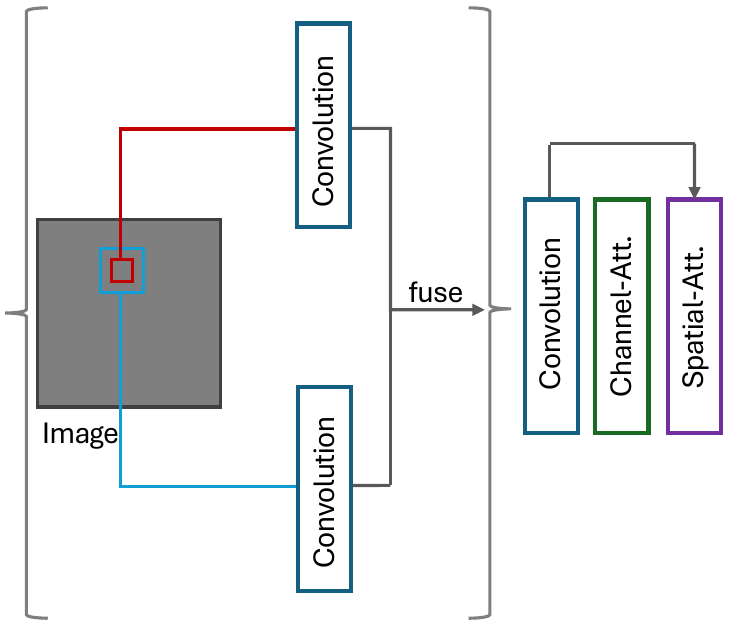}
  \caption{Multiscale convolution block used in the modality autoencoder. 
  Local and broader spatial context are captured through parallel convolutional paths; 
  features are fused by convolution and channel-attention before being passed to the temporal encoder (Fig.~\ref{fig:modality-encoder}).}
  \label{fig:multiscale_block}
\end{figure}

Each autoencoder processes these spatio-temporal patches to learn the intrinsic structures of radar or optical observations that encode spatial and temporal context. The architecture, illustrated in Fig.~\ref{fig:modality-encoder}, follows an encoder–decoder framework that integrates convolutional attention mechanisms to refine feature extraction. It contains three convolutional attention blocks in both the encoder and decoder components (Fig.~\ref{fig:multiscale_block}). Each block is equipped with a Convolutional Block Attention Module (CBAM)~\cite{woo2018cbam}, which applies channel and spatial attention to emphasize spectrally and structurally relevant features. The first encoder and final decoder blocks additionally include a multiscale module~\cite{szegedy2015going} (gray brackets in Fig.~\ref{fig:multiscale_block}) that aggregates spatial information through parallel convolutions with different kernel sizes, enabling the model to jointly capture fine spatial details and broader contextual patterns. Temporal dependencies across the 11 input frames are modeled by a transformer encoder~\cite{vaswani2017attention} that uses positional encodings derived from actual irregular acquisition intervals, allowing the network to represent both short-term variability and seasonal dynamics.

The latent bottleneck includes two dimensions for \mbox{Sentinel-1} and nine for \mbox{Sentinel-2}. It is designed to internalize the central pixel for each modality as accurately as possible and encode its environmental context. The decoder mirrors the encoder to reconstruct the full input sequence from these latent representations. Training is guided by a context-aware loss that assigns the highest weight to the central coordinate while progressively down-weighting surrounding pixels and timesteps (Fig.~\ref{fig:context-aware-MAE}). This strategy encourages accurate reconstruction of the target location while ensuring that the learned embeddings remain informed by their spatial and temporal neighborhood.

Since the context-aware learning strategy assigns varying weights to the individual loss components, prioritising the central pixel, each component is analysed separately to assess how well the model captures both local and contextual information.

\subsection{Data Fusion}
\label{data-fusion}

\begin{figure}[t]
  \centering
  \includegraphics[width=\linewidth]{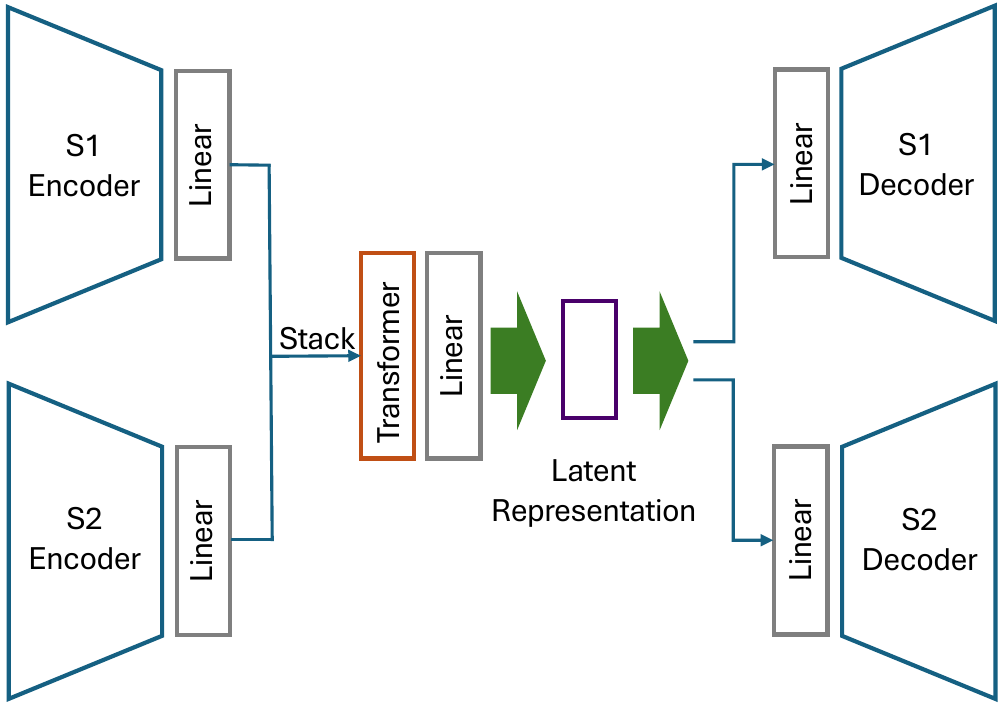}
  \caption{Multimodal data-fusion architecture combining pretrained \mbox{Sentinel-1} and \mbox{Sentinel-2} autoencoders.
Latent features from each modality are projected and stacked, processed by a Transformer with temporal positional encodings, and mapped into a shared latent representation.}
  \label{fig:modality_fusion}
\end{figure}

After modality-specific pretraining, the two autoencoders were combined into a joint fusion network (Fig.~\ref{fig:modality_fusion}).
The main objective of this stage is to learn a compact and consistent embedding with increased emphasis on the central coordinate of each spatio-temporal patch by integrating complementary radar and optical information.

For fusion pretraining, \mbox{Sentinel-1} observations were temporally matched to the closest \mbox{Sentinel-2} acquisitions to ensure aligned radar–optical pairs.
As in the pretraining phase, samples were extracted as spatio-temporal patches of $15\times15$ pixels, now containing 12 spectral channels in total (two from \mbox{Sentinel-1} and ten from \mbox{Sentinel-2}).
From each 20-frame temporal sequence, 11 frames were randomly selected to improve temporal generalization. A temporal stride of 17 and a spatial stride of 9 were applied across the aligned modalities to increase diversity and spatial coverage.

During modality fusion, the pretrained encoders and decoders remained frozen, while only the newly introduced fusion layers were optimized.
Latent features from the \mbox{Sentinel-1} and \mbox{Sentinel-2} encoders were first passed through modality-specific linear projection layers and then concatenated.
The combined features were processed by a Transformer encoder equipped with temporal positional encodings, enabling the model to account for asynchronous radar–optical acquisitions and to capture cross-modal temporal dependencies.
Subsequently, a linear projection layer produced a compact seven-dimensional bottleneck representation encoding the central pixel within a temporally consistent, multimodal embedding space.
The bottleneck dimensionality was chosen empirically as it achieved the lowest reconstruction MAE for the central pixel of the spatio-temporal patch.
These fused representations were then projected back through linear layers and reconstructed by the respective decoders.

Validation during modality fusion focused exclusively on the reconstruction accuracy of the central pixel, as this coordinate represents the primary target of the context-aware learning framework. Accordingly, the loss function was specifically designed to minimize the validation MAE of this central point as effectively as possible while including contextual contributions from its spatial and temporal neighborhood as additional cues for an improved reconstruction.
For \mbox{Sentinel-1}, again the weighted MAE loss was used as the training loss. For this fusion phase we increased the central pixel weight to 1.5 to emphasize its reconstruction accuracy, while the remaining weights were kept identical to those used during pretraining.
For \mbox{Sentinel-2}, this weight was set to 1.75 due to its higher reconstruction error in pretraining. Additionally, SAM was applied to the central coordinate in the \mbox{Sentinel-2} loss to further enforce spectral consistency at the reconstruction target.
The resulting \mbox{Sentinel-2} loss function was defined as:
\begin{equation}
\mathcal{L}_{\text{S2}} = 0.85,\mathcal{L}_{\text{MAE}} + 0.15,\mathcal{L}_{\text{SAM}}.
\end{equation}

To promote structural and spectral coherence between the two modalities, a joint fusion loss was introduced, combining the SSIM loss and SAM:
\begin{equation}
\mathcal{L}_{\text{joint}} = 0.1,\mathcal{L}_{\text{SSIM}} + 0.9,\mathcal{L}_{\text{SAM}}.
\end{equation}

The total fusion objective combined modality-specific and joint losses as
\begin{equation}
\mathcal{L}_{\text{total}} = 0.45,\mathcal{L}_{\text{S1}} + 0.45,\mathcal{L}_{\text{S2}} + 0.10,\mathcal{L}_{\text{joint}},
\end{equation}
balancing accurate modality reconstruction with cross-modal alignment.

This configuration encouraged the network to form a coherent joint embedding that preserves the structural and spectral integrity of radar and optical observations while achieving accurate reconstruction of the central coordinate.

\begin{figure*}[ht]  
  \centering
  \includegraphics[page=1,width=\textwidth]{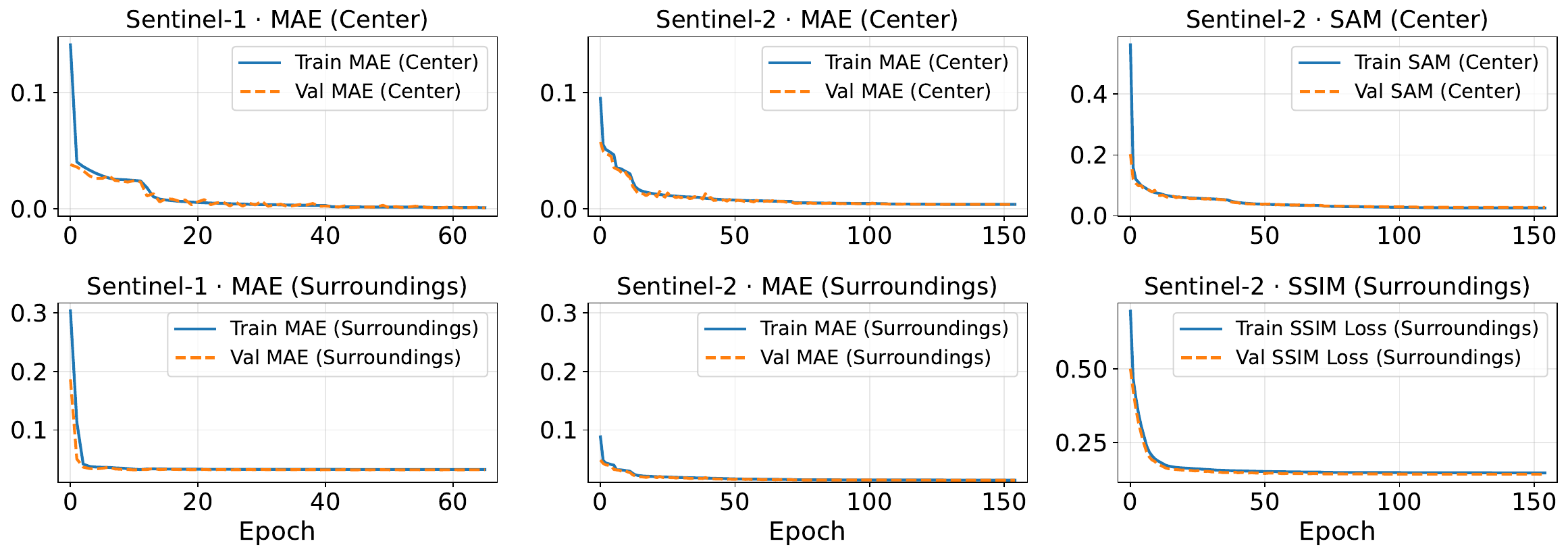}
  \caption{Training and validation losses for \mbox{Sentinel-1} (left column) and \mbox{Sentinel-2} (middle and right columns) models during context-aware pretraining.
The first row shows losses computed for the central pixel only, including MAE and SAM, while the second row depicts corresponding metrics for the spatial surroundings (MAE and SSIM).
Both models exhibit stable convergence and consistent performance between training and validation, indicating effective reconstruction of both central and contextual features.}
  \label{fig:loss-components-pretraining}
\end{figure*}

\section{Evaluation \& Results}
\label{sec:evaluation}

To confirm stable convergence and generalization across modalities and fusion stages, we first analyse the training and validation behavior of the model (Section~\ref{training-validation}). 
Subsequently, we examine the spatial organization of the fused feature space by visualising its principal components and comparing them to corresponding \mbox{Sentinel-2} RGB imagery, assessing the spatial coherence and semantic consistency of the learned representations. 
Finally, we evaluate the temporal expressiveness of the embeddings in a downstream Gross Primary Productivity (GPP) modelling task, testing their capacity to represent vegetation dynamics and ecosystem functioning across multiple sites and vegetation types. 
Together, these analyses provide a comprehensive assessment of the framework’s performance during training and demonstrate its ability to generate analysis-ready latent feature datasets with high spatial and temporal resolution from multiple modalities.

\subsection{Model Training and Validation}
\label{training-validation}

This section evaluates the training behavior and validation performance of the proposed framework. 
We first analyse the modality-specific pretraining stage, where independent autoencoders for \mbox{Sentinel-1} and \mbox{Sentinel-2} learn to reconstruct radar and optical observations in a context-aware manner. 
Subsequently, we assess the multimodal fusion stage, in which the pretrained encoders are combined and fine-tuned through lightweight fusion layers to produce coherent joint embeddings. 
Quantitative metrics and loss evolution curves are presented to demonstrate the stability, convergence, and generalization capability of both training stages.

\subsubsection{Modality Pretraining Results}
\label{pretraining-results}

Since the context-aware learning strategy assigns varying weights to the individual loss components, prioritising the central pixel, each component was analysed separately to assess how well the model captures both local and contextual information. 
As shown in Fig.~\ref{fig:loss-components-pretraining}, the upper row displays the evolution of losses for the central pixel (MAE and SAM), while the lower row presents the corresponding metrics for the surrounding spatial context (MAE and SSIM) for both \mbox{Sentinel-1} and \mbox{Sentinel-2} models. 
This separation illustrates how the model learns to reconstruct the target location with high precision while simultaneously integrating meaningful information from neighboring pixels and time frames. 
Both modality-specific autoencoders show smooth convergence and close agreement between training and validation curves, confirming that the context-weighted objective ensures stable optimization without overfitting.

Quantitatively, the weighted MAE for \mbox{Sentinel-1} converged to 0.0009 (training) and 0.0006 (validation) for the central pixel, and to 0.0324 (training) and 0.0319 (validation) for the surrounding context. 
For \mbox{Sentinel-2}, the central-pixel MAE reached 0.0037 (training) and 0.0037 (validation), while SAM converged to 0.0259\,rad ($\approx$1.5$^\circ$) in training and 0.0278\,rad ($\approx$1.6$^\circ$) in validation, confirming minimal angular deviation between reconstructed and reference spectra. 
The SSIM loss stabilized at 0.1476 (training) and 0.1431 (validation), and the contextual weighted MAE reached 0.0142 (training) and 0.0141 (validation). 
Overall, the close correspondence between training and validation metrics across all loss components highlights the generalization capability of the autoencoders, reconstructing the target pixel while embedding surrounding context within each modality.

\subsubsection{Modality Fusion}

The goal of this stage was to learn a compact and consistent embedding with increased emphasis on the central coordinate of each spatio-temporal patch by integrating complementary radar and optical information.
To achieve this, we designed a training loss that minimizes the MAE of the central point within each spatio-temporal patch as efficiently as possible (Section~\ref{data-fusion}). 
Validation during modality fusion focused on the central-pixel reconstruction accuracy, as this coordinate represents the primary target of the context-aware learning framework (Section~\ref{context-aware-learning}).

The final fusion model achieved a validation MAE of 0.0004 for \mbox{Sentinel-1} and 0.0039 for \mbox{Sentinel-2}, suggesting accurate reconstruction across both modalities.
The test set confirmed these results, yielding nearly identical MAE values of 0.0004 and 0.0040, respectively, indicating stable generalization.

\subsection{Qualitative Evaluation of the Learned Feature Space}

\begin{figure*}[t]
    \centering
    \includegraphics[width=\textwidth,keepaspectratio]{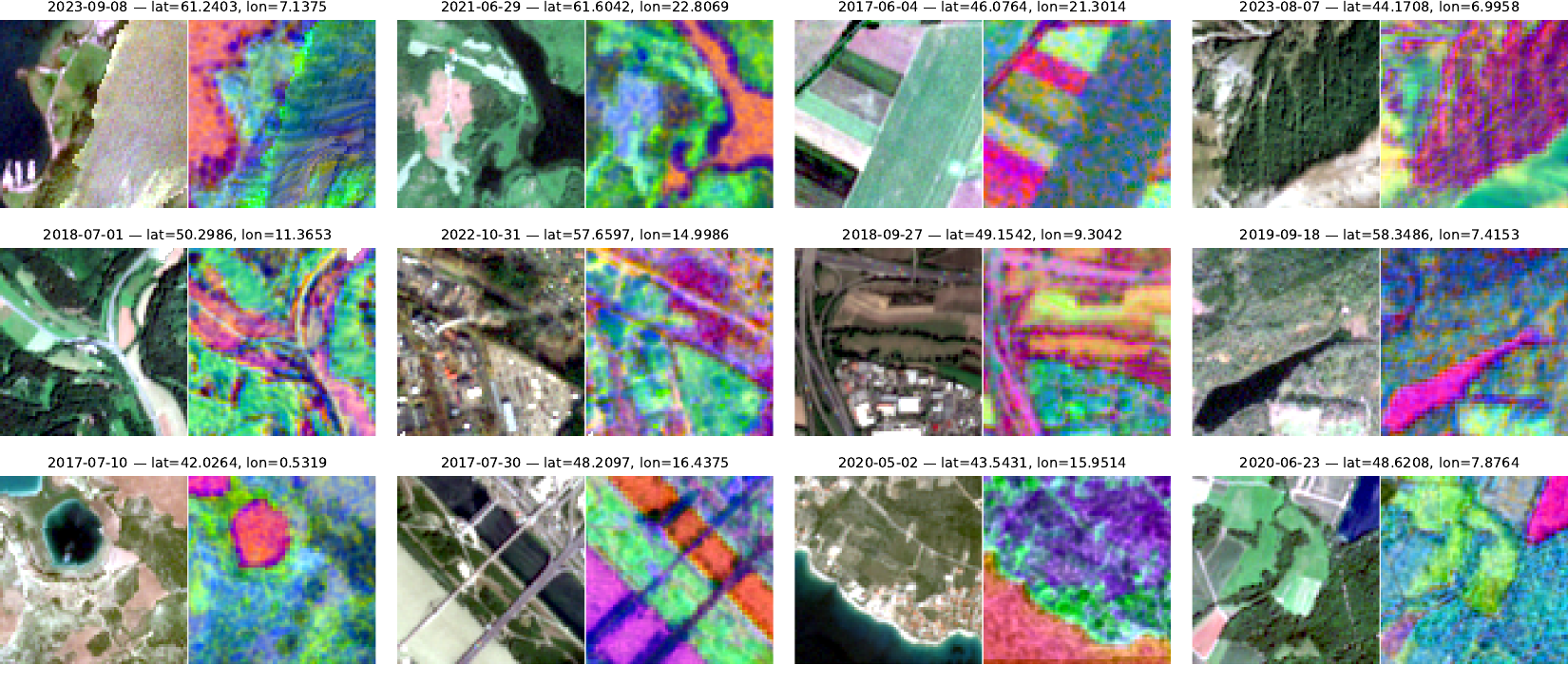}
    \caption{
Comparison between \mbox{Sentinel-2} RGB imagery (left in each pair) and PCA projections of the learned feature embeddings (right) for twelve representative locations.
The left two columns show examples from the training set, while the right two columns depict samples from the validation and test sets.
Each pair corresponds to the same date and spatial extent, illustrating the semantic richness and spatial consistency of the learned feature space across different regions and data splits.
}
    \label{fig:qualitative_features}

\end{figure*}

To qualitatively assess the structure of the learned multimodal feature space, we visualise the first three principal components of the fused \mbox{Sentinel-1} and \mbox{Sentinel-2} embeddings and compare them to the corresponding \mbox{Sentinel-2} RGB observations (Figure~\ref{fig:qualitative_features}).
Each pair represents the same spatial extent and acquisition date, providing an intuitive comparison between the raw optical signal and the internal representation learned from radar and optical inputs.
The selected dates correspond to scenes with the highest data availability across both sensors.

The PCA projections reveal coherent and interpretable spatial patterns that align closely with the land-surface structures visible in the RGB reference imagery.
Distinct color regions delineate vegetation, bare soil, water, and urban areas, indicating that the learned embeddings separate major land-cover types in a semantically meaningful way.
The overall spatial organization remains consistent across diverse landscapes and data splits, providing evidence that the fused \mbox{Sentinel-1} and \mbox{Sentinel-2} embeddings capture stable and interpretable spatial representations that generalize beyond the training regions.

\subsection{GPP Modelling}
\label{sec:gpp_modelling}
Accurate quantification of GPP is essential for understanding terrestrial carbon dynamics, yet direct measurements from Eddy Covariance (EC) towers remain spatially sparse. 
Montero~\textit{et al.}~\cite{montero2024recurrent} addressed this limitation by developing a multimodal deep learning framework that estimates daily GPP from satellite time series across 19 forest sites.
Their model combined \mbox{Sentinel-1} radar data, \mbox{Sentinel-2} optical reflectance, MODIS  land surface temperature, and solar radiation inputs within a recurrent neural network to capture the temporal evolution of photosynthetic activity.
Building on this concept, we evaluated our learned \mbox{Sentinel-1} and \mbox{Sentinel-2} embeddings using a Transformer-based regression framework.
The model architecture consists of a linear input layer, four Transformer encoder blocks,
and a regression head producing a single daily GPP estimate. 

We assembled daily, 10\,m \mbox{Sentinel-1} and \mbox{Sentinel-2} embeddings over multiple FLUXNET sites in Central Europe for the 2017--2020 period and linearly interpolated missing observations to ensure continuous temporal coverage. 
Daily GPP values (g\,C\,m$^{-2}$\,d$^{-1}$) were obtained from the nighttime partitioning method (\texttt{GPP\_NT\_VUT\_REF})~\cite{reichstein2005separation}. 
Timesteps with less than 70\,\% high-quality measurements (based on \texttt{NEE\_VUT\_REF\_QC}) and negative GPP values were excluded from training to ensure robust target data. 
We selected sites for which more than 60\,\% of the remaining GPP observations passed the 70\,\% quality criterion, prioritising those with high temporal completeness and minimal data gaps across three vegetation types. 
The final site set includes nine locations representing deciduous broadleaf forest (DBF: CZ-Lnz, DE-HoH, FR-Hes), evergreen needleleaf forest (ENF: BE-Bra, IT-Lav, IT-Ren), and grassland (GRA: CH-Cha, CH-Fru, IT-MBo).

\begin{figure}[ht]
    \centering
    \includegraphics[width=\linewidth]{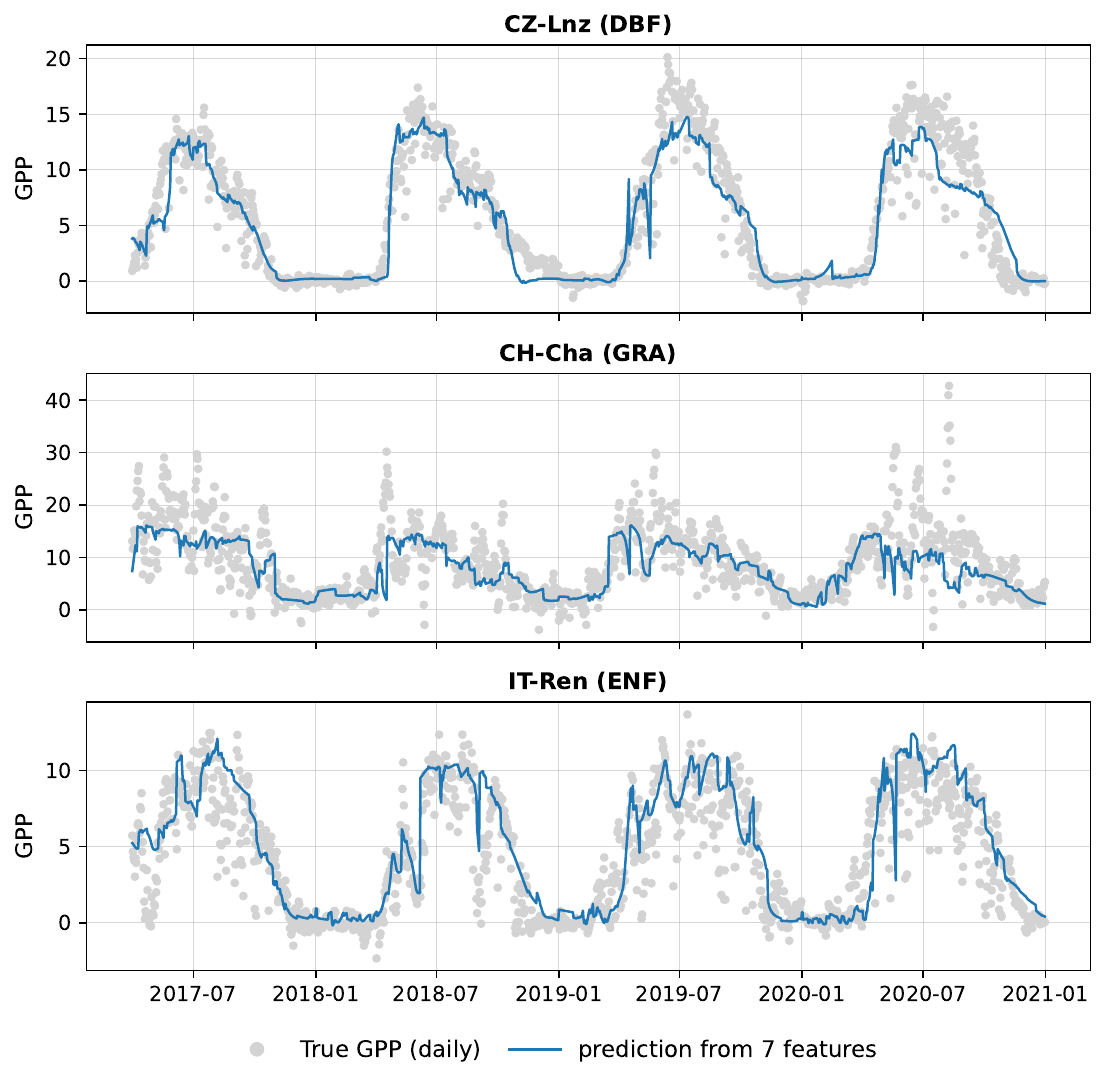}
    \caption{
    Daily GPP reconstructions from \mbox{Sentinel-1} and \mbox{Sentinel-2} embeddings across three vegetation types (DBF/ENF/GRA). 
    }
    \label{fig:gpp_compare}
\end{figure}

We systematically tested temporal sequence lengths between 60 and 120 days to determine the optimal input context for predicting daily GPP. To increase the number of available samples while maintaining temporal coherence, we sampled all training and validation sequences with a stride of 10 days. A 90-day window provided the best trade-off between predictive accuracy and temporal stability across sites, consistent with the findings of~\cite{montero2024recurrent}. Following this setup, we trained our model to predict daily GPP from 90-day input sequences.
We defined training and validation splits by year, using 2017–2019 for training and 2020 for validation. We optimized the model using the MAE loss function and evaluated performance with the normalized root mean square error (NRMSE), achieving 0.0917 on the training and 0.0958 on the validation set.
In addition, the global RMSE across all sites reached 3.09 for training and 4.40 for validation.

Montero~\textit{et al.}~\cite{montero2024recurrent} reported a validation NRMSE of about 0.13 when predicting GPP across 19 forest sites using multimodal observations.
While our results show lower errors, the setups are not directly comparable, as we train on fewer sites and slightly different vegetation types.
Another related study \cite{pabon2022potential} estimated GPP based on \mbox{Sentinel-2} spectral indices and bands from 58 EC sites. 
Their machine-learning approach achieved RMSE values be tween 2.5 and 3.0. 
These results indicate that our learned multimodal embeddings effectively capture relevant information for predicting carbon fluxes, providing performance comparable to models to previous approaches.

As illustrated in Figure~\ref{fig:gpp_compare}, our 7-feature model captures site-specific productivity patterns across all vegetation types. The results across the three ecosystems suggest that the learned \mbox{Sentinel-1} and \mbox{Sentinel-2} embeddings encode ecologically meaningful information related to vegetation productivity and its temporal variability.

\section{Discussion}
The presented data fusion framework provides a flexible and analysis-ready foundation for environmental modelling by integrating complementary information from multiple Earth observation modalities.
Through multimodal, context-aware representation learning, the framework integrates heterogeneous EO data into a unified latent feature space at high spatio-temporal resolution.
We demonstrate the approach using \mbox{Sentinel-1} radar and \mbox{Sentinel-2} optical data as representative modalities.
    
The resulting representations enable flexible analyses for a broad range of downstream applications that require different spatial and temporal contexts, from short-term vegetation dynamics to long-term ecosystem trends.
Because the embeddings retain explicit spatio-temporal structure at 10m resolution and follow the natural acquisition frequency of the sensors after cloud masking, they can be aggregated or sampled flexibly to match diverse ecological and climate-modelling requirements.

In our context-aware loss function, we chose $\alpha = 0.1$ to prioritise the immediate neighborhood while still preserving broader contextual cues.
Although this choice yielded stable training and consistent results across modalities, we did not experiment with different $\alpha$ values, leaving a systematic exploration of its sensitivity for future work.
More generally, other hyperparameters, such as the \mbox{Sentinel-2} reconstruction loss weights during pretraining or the fusion loss configuration were determined through limited empirical experimentation rather than extensive tuning.
We did not perform a comprehensive hyperparameter optimization due to hardware constraints.
Nevertheless, the framework demonstrated stable convergence and reliable reconstruction across modalities.

A qualitative evaluation confirmed that the learned feature embeddings exhibit strong spatial and semantic consistency across heterogeneous landscapes.  
When projected into three dimensions via PCA and compared with corresponding \mbox{Sentinel-2} RGB scenes (Fig.~\ref{fig:qualitative_features}), the embeddings preserved land-surface patterns and gradients associated with vegetation, water, and soil characteristics, revealing that the model effectively captures physically meaningful information shared across modalities.  
This provides visual evidence that the learned features encode interpretable environmental structure rather than abstract or sensor-specific noise.

Using the learned multimodal embeddings, the model successfully reproduced site-specific GPP dynamics with accuracy comparable to previous multispectral and multimodal approaches~\cite{pabon2022potential, montero2024recurrent}.
These results indicate that the representations capture information relevant for modelling photosynthetic activity across diverse vegetation types and environmental conditions.
The ability to predict daily GPP consistently across different sites demonstrates that the fused feature space encodes meaningful temporal signals that relate to ecosystem functioning.

The agreement between the fused embeddings and GPP dynamics indicates that the model has learned physically grounded representations of the land surface. \mbox{Sentinel-2} contributes spectral signals related to canopy chlorophyll, vegetation density, and leaf water content through its near-infrared and short-wave infrared bands~\cite{ceccato2001detecting, jackson2004vegetation}. \mbox{Sentinel-1} complements this with sensitivity to vegetation structure, biomass, and surface moisture owing to the dielectric and geometric response of C-band microwaves~\cite{dubois2020characterization, frison2018potential}.
Together, these modalities encode biophysical variability closely associated with seasonal productivity patterns across sites.

A major advantage of the learned embeddings is their high temporal and spatial flexibility.
Users can aggregate them over larger spatial units or select task-specific temporal windows depending on the application.
This flexibility enabled the 90-day sliding-window setup used in our GPP experiment, which captures short-term seasonal dynamics while preserving 10 m spatial detail.

In contrast, yearly embeddings require predicting all daily GPP values simultaneously from a single latent representation. This provides too few samples per site to learn robust temporal relationships.
Previous work has shown that the EC footprint varies substantially across seasons and sites. Consequently, approaches that rely on coarse spatial resolution can introduce representativeness biases, reducing the reliability of GPP analyses \cite{schmid1997experimental, chen2009assessing, chu2021representativeness}. High spatial resolution provides the flexibility needed to select footprint-aligned areas, ensuring that the domain used for model assessment closely matches the conditions actually observed by the tower \cite{chu2021representativeness}. Our spatio-temporally explicit approach directly addresses both limitations.

A current limitation is that the learned features contain temporal and spatial gaps, as missing acquisitions from \mbox{Sentinel-1} or cloud-filtered \mbox{Sentinel-2} propagate into the feature space.  
In some periods, especially during winter or persistent cloud cover, these gaps can span days to weeks.  
Established gap-filling techniques offer practical ways to mitigate such effects.
These include spatial interpolation approaches such as kriging \cite{addink1999comparison, rossi1994kriging} or temporal smoothing methods including the Savitzky–Golay and seasonal filters implemented in TIMESAT \cite{jonsson2004timesat}, and hybrid spatio-temporal strategies that alternate between spatial and temporal modelling steps \cite{kang2005improving, borak2009effective}.
Alternatively, machine learning methods can be applied~\cite{weiss2014effective}.

To simplify the integration of additional sensors, our approach first models each modality independently and trains only the fusion layers jointly. This design allows every encoder to be optimised with modality-specific learning strategies while keeping the overall framework flexible and extensible.
A remaining challenge arises from differences in spatial resolution. Variations in ground sampling and geometric accuracy can introduce scale-dependent inconsistencies and spatial artefacts when combining heterogeneous data~\cite{schowengerdt2006remote, guo2024obsum, wang2021blocks}. These effects may reduce spatial coherence, particularly in heterogeneous landscapes. However, scale-aware processing strategies, such as adaptive resampling~\cite{arun2014intelligent} or attention-based fusion mechanisms~\cite{gao2023cross}, provide effective ways to mitigate these issues and support consistent integration across resolutions.

Overall, our framework contributes significantly to the development of general-purpose representations for environmental modelling.  
Both the qualitative analysis and the predictive experiments confirm that the learned \mbox{Sentinel-1} and \mbox{Sentinel-2} embeddings encode ecologically meaningful information, capturing spatial structure, surface characteristics, and seasonal vegetation dynamics across diverse landscapes.  
This demonstrates that the learned feature space provides a transferable, flexible, and scalable foundation for data-driven Earth system science, capable of supporting a wide range of downstream applications that require various spatial and temporal context of high resolution.

\section{Conclusion}
In this work, we presented a multimodal, context-aware representation learning framework that integrates \mbox{Sentinel-1} radar and \mbox{Sentinel-2} (10–20 m) optical observations into a unified, spatio-temporally explicit feature space of high-resolution. By combining staged pretraining, context-aware reconstruction, and transformer-based fusion, the framework effectively captures complementary spectral and structural information while preserving spatial coherence and the temporal fidelity of \mbox{Sentinel-2} acquisitions.

Our primary objective was to create analysis-ready, spatio-temporally explicit embeddings that can flexibly support diverse environmental modelling tasks.
The results confirm that the learned representations generalize across both modalities, achieving high reconstruction accuracy and stable validation performance.
Qualitative evaluation demonstrated that the fused feature space preserves land-surface structures and semantic consistency across diverse regions, while quantitative experiments on GPP modelling confirmed the relevance of the learned representations for flexible environmental modelling.
Together, these findings demonstrate that the proposed framework contributes to bridging large-scale representation learning with fine-scale environmental modelling, producing transferable embeddings that retain detailed spatio-temporal structure.

Future work can focus on extending the framework toward broader multi-sensor integration and enhanced temporal continuity.
Promising directions include addressing data gaps through cross-modal learning or spatio-temporal interpolation, and developing scale-aware fusion strategies to harmonize sensors with different resolutions.
Such advancements would further strengthen the adaptability and ecological interpretability of the learned representations, supporting more comprehensive, data-driven analyses of Earth system dynamics across scales and sensor constellations.


%



\section*{Acknowledgment}
KM, JP, CM, MR, and MDM acknowledge funding by the European Space Agency AI4Science project `DeepFeatures' (2024--2026).
CM, MR, and MDM thank the Bundesministerium für Wirtschaft und Klimaschutz for funding ML4Earth, Grant Number: 50EE2201B.

\ifCLASSOPTIONcaptionsoff
  \newpage
\fi



%
%
\bibliographystyle{IEEEtran}
\bibliography{bibtex/bib/IEEEexample}









\end{document}